\documentclass{article}

\usepackage[final]{neurips_2024}

\usepackage[numbers]{natbib}
\usepackage[utf8]{inputenc} 
\usepackage[T1]{fontenc}    
\usepackage{hyperref}       
\usepackage{url}            
\usepackage{booktabs}       
\usepackage{amsfonts}       
\usepackage{nicefrac}       
\usepackage{microtype}      
\usepackage{xcolor}         
\usepackage{sidecap}
\usepackage{multirow}
\usepackage{tabularx}
\usepackage{colortbl}
\usepackage{graphicx}
\usepackage{booktabs}
\usepackage{array}
\usepackage{arydshln}
\usepackage{amsmath}
\usepackage{lipsum}
\usepackage{amsmath}
\usepackage{tikz}
\usepackage{wrapfig}
\usepackage{enumitem}
\usepackage{xspace}

\usepackage{soul}
\usepackage{xcolor}

\definecolor{pink}{HTML}{A02993}
\definecolor{blue}{HTML}{4E95D9}
\definecolor{green}{HTML}{459528}

\newcommand{\ourmodel}{\xspace\texttt{VideoGPT+}\xspace}
\newcommand{\ourmodelbold}{\xspace\textbf{{VideoGPT+}}\xspace}
\newcommand{\ourdata}{\xspace\texttt{VCG+\,112K}\xspace}
\newcommand{\ourdatabold}{\xspace\textbf{{VCG+\,112K}}\xspace}
\newcommand{\ourbench}{\xspace\texttt{VCGBench-Diverse}\xspace}
\newcommand{\ourbenchbold}{\xspace\textbf{{VCGBench-Diverse}}\xspace}

\newcommand{\rom}[1]{{(\romannumeral #1)}}

\title{VideoGPT+ \includegraphics[width=0.06\linewidth]{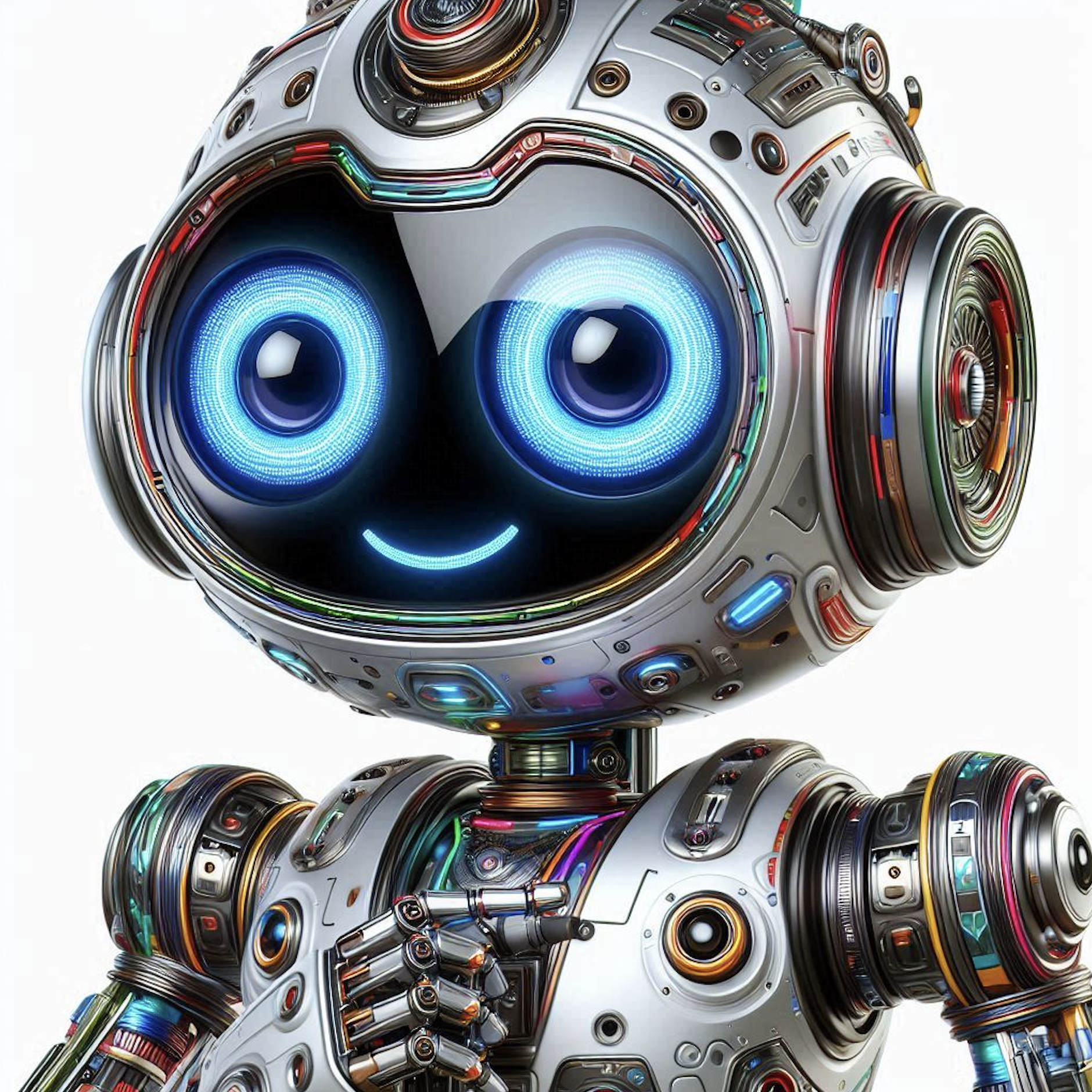}: Integrating Image and Video Encoders for Enhanced Video Understanding}

\author{%
Muhammad Maaz\textsuperscript{\textnormal{1}},
  Hanoona Rasheed\textsuperscript{\textnormal{1}}, 
  \textbf{Salman Khan}\textsuperscript{\textnormal{1,2}},
  \textbf{Fahad Shahbaz Khan}\textsuperscript{\textnormal{1,3}} \\
  \textsuperscript{1}Mohamed bin Zayed University of AI, UAE \\ 
  \textsuperscript{2}Australian National University, Australia \quad \textsuperscript{3}Linköping University, Sweden
    }

\begin{document}

\maketitle

\begin{abstract}
Building on the advances of language models, Large Multimodal Models (LMMs) have contributed significant improvements in video understanding. 
While the current video LMMs utilize advanced Large Language Models (LLMs), they rely on either image or video encoders to process visual inputs, each of which has its own limitations. 
Image encoders excel at capturing rich spatial details from frame sequences but lack explicit temporal context, which can be important in videos with intricate action sequences.
On the other hand, video encoders provide temporal context but are often limited by computational constraints that lead to processing only sparse frames at lower resolutions, resulting in reduced contextual and spatial understanding.
To this end, we introduce \ourmodel, which combines the complementary benefits of the image encoder (for detailed spatial understanding) and the video encoder (for global temporal context modeling). 
The model processes videos by dividing them into smaller segments and applies an adaptive pooling strategy on features extracted by both image and video encoders. 
Our architecture showcases improved performance across multiple video benchmarks, including VCGBench, MVBench and Zero-shot question-answering. 
Further, we develop 112K video-instruction set using a novel semi-automatic annotation pipeline which further improves the model performance. 
Additionally, to comprehensively evaluate video LMMs, we present \ourbench, covering 18 broad video categories such as lifestyle, sports, science, gaming, and surveillance videos. 
This benchmark with 4,354 question-answer pairs evaluates the generalization of existing LMMs on dense video captioning, spatial and temporal understanding, and complex reasoning, ensuring comprehensive assessment across diverse video types and dynamics. Code: \url{https://github.com/mbzuai-oryx/VideoGPT-plus}.
\vspace{-0.2cm}
\end{abstract}

\section{Introduction}
Existing methods for video understanding often rely solely on either image encoders or video encoders~\cite{Maaz2023VideoChatGPT,jin2023chatunivi, st-llm}. Most works focus on image encoders, which encode multiple frames and either fuse the information or concatenate the embeddings before passing them to the LLM. When fusing the information, spatial or temporal pooling is typically used~\cite{Maaz2023VideoChatGPT}. Spatial pooling has shown minimal effectiveness in capturing video information, whereas temporal pooling retains some spatial information but lacks explicit temporal context. On the other hand, concatenating embeddings without pooling~\cite{jin2023chatunivi, st-llm, zhang2024llavanextvideo} can rapidly increase computational complexity due to the extended context length required by the LLM, limiting the number of frames that can be processed. While this approach provides better spatial representation, the overall context is still limited to few frames. The limited context results in a poor understanding of the video, especially if a uniform sampling strategy is employed, as it only captures small segments of the video, missing important temporal dynamics.

\begin{wrapfigure}{r}{0.5\linewidth}
  \centering
  \includegraphics[width=\linewidth]{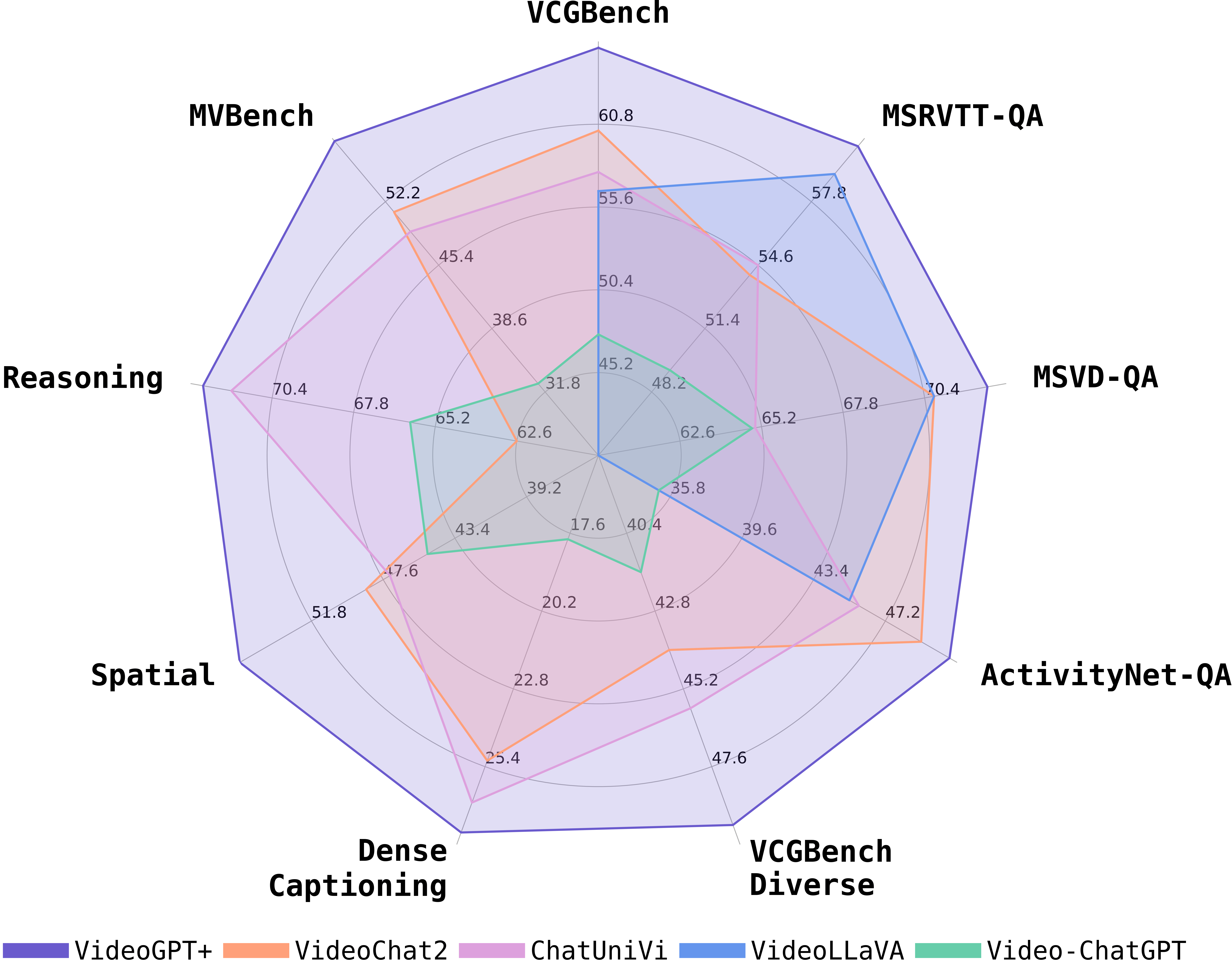}
  \vspace{-1.6em}
  \caption{\small \textbf{Performance comparison of \ourmodelbold} with various SoTA models across multiple video benchmarks. \ourmodel~demonstrates better performance compared to various models~\cite{2023videochat, jin2023chatunivi, video-llava, Maaz2023VideoChatGPT}
  on video conversation benchmarks: VCGBench~\cite{Maaz2023VideoChatGPT} and MVBench~\cite{2023videochat}, Zero-shot video question answering: MSVD-QA, MSRVTT-QA, ActivityNet-QA. We also evaluate on \ourbench~that covers 18 broad video categories (across dense captioning, spatial understanding, and reasoning).}
  \label{fig:intro_block_diagram}
  \vspace{-1.6em}
\end{wrapfigure}
In order to address these challenges, we propose \ourmodel which effectively combines the merits of both image and video encoders (see Fig.~\ref{fig:arch_block_diagram}). By leveraging an image encoder for rich spatial details and a video encoder for global temporal context, our model achieves improved video understanding. To model finegrained temporal dynamics in \ourmodel~, we use a segment-wise sampling strategy. Unlike uniform sampling used in existing video LMMs \cite{Maaz2023VideoChatGPT}, which may miss important temporal dynamics, our approach divides the video into smaller segments and applies segment-wise sampling. This ensures that the model captures representative information from different segments of the video, enabling a more 
comprehensive understanding.

To facilitate the integration of image and video features, \ourmodel~introduces a visual adapter module that combines their complimentary benefits. This module performs projection and pooling operations, mapping both image and video features to a common space while reducing computational complexity. By aligning the features in this manner, the model can effectively utilize the combined spatial and temporal information for improved video understanding.

We demonstrate the effectiveness of \ourmodel~across multiple video-conversation benchmarks, including VCGBench \cite{Maaz2023VideoChatGPT}, MVBench \cite{li2023mvbench}, and Zero-shot question-answering \cite{Maaz2023VideoChatGPT}, where it outperforms previous SoTA approaches (see Fig.~\ref{fig:intro_block_diagram}). Further, we develop \ourdata~using a novel semi-automatic annotation pipeline (see Fig.~\ref{fig:data_block_diagram}), which provides dense video captions along with spatial understanding and reasoning-based question-answer (QA) pairs, further enhancing the model's performance. We also propose \ourbench, extending VCGBench~\cite{Maaz2023VideoChatGPT} by including videos from 18 different domains to extensively evaluate the video-based conversation models in diverse domains (see Fig.~\ref{fig:benchmark_block_diagram}).

Our work has three main contributions:
\vspace{-0.7em}
\begin{itemize}[leftmargin=1em,itemsep=0mm]
    \item We present~\ourmodel, the first video-conversation model that benefits from a dual-encoding scheme based on both image and video features. These complimentary sets of features offer rich spatiotemporal details for improved video understanding (Sec.~\ref{method}).
    \item Addressing the limitations of existing VideoInstruct100K dataset~\cite{Maaz2023VideoChatGPT}, we develop \ourdata with a novel semi-automatic annotation pipeline, offering dense video captions along with spatial understanding and reasoning-based QA pairs, further improving the model performance (Sec.~\ref{dataset}).
    \item Recognizing the lack of diverse benchmarks for video-conversation task, we propose \ourbench, which provides 4,354 human annotated QA pairs across 18 video categories to extensively evaluate the performance of a video-conversation model (Sec.~\ref{benchmarks}).
\end{itemize}

\section{Related Works}
Building on advances in language models, LLMs offer a flexible interface for various multimodal applications. Early efforts in image-based conversation models such as BLIP-2~\cite{li2023blip}, MiniGPT-4~\cite{zhu2023minigpt} and LLaVA~\cite{liu2023llava, liu2023improvedllava} project image features into the language space through a learnable module and perform instruction tuning for visual conversations capabilities. Other efforts extend these models to visual grounding tasks~\cite{kosmos-2, hanoona2023GLaMM, you2023ferret}, exploring the potential of LLMs in complex vision tasks.

\textbf{Video Conversation Models:} Initial works like Video-ChatGPT~\cite{Maaz2023VideoChatGPT} and Video-LLaMA~\cite{damonlpsg2023videollama} extend image-based models to the video domain by introducing components to encode temporal features, where frame-level visual features are fed to the LLM. However, this is computationally expensive and quickly fills its context window. To address this issue, Video-ChatGPT~\cite{Maaz2023VideoChatGPT} employs spatial and temporal pooling. LLaMA-Vid~\cite{llamavid} proposes representing a single image with two tokens, context and content. IG-VLM~\cite{kim2024image} treats a video as a grid of images, while LITA~\cite{huang2024lita} employs slow-fast token pooling to reduce the number of visual features. Chat-UniVi~\cite{jin2023chatunivi} uses clustering in both spatial and temporal dimensions to merge tokens, and VideoChat~\cite{2023videochat} uses Q-Former~\cite{li2023blip} to learn a fixed number of queries by cross-attending to the visual features. MobileVLM~\cite{chu2023mobilevlm, chu2024mobilevlm} utilize a lightweight CNN to reduce the spatial dimensions. Other notable methods include \cite{bt_adapter,video-llava,munasinghe2023PGVideoLLaVA,song2023moviechat,huang2023vtimellm}.

Alternatively, methods such as VideoChat2~\cite{li2023mvbench} use pretrained video encoders. Although video encoders provide temporal context, they are limited by computational constraints, operating with limited frames at lower resolutions, restricting temporal context and spatial understanding. Our \ourmodel model addresses these issues by using segment-wise sampling and effectively combining the merits of image and video encoders to capture rich 
spatial and temporal details (see Fig.~\ref{fig:arch_block_diagram}).

\textbf{Video Instruction Tuning Datasets:} VideoChat~\cite{2023videochat} builds a video-instruction tuning dataset consisting of 7K instructions using videos from WebVid-10M~\cite{bain2021frozen}. Video-ChatGPT~\cite{Maaz2023VideoChatGPT} introduces a semi-automatic annotation pipeline to generate VideoInstruct100K using videos from ActivityNet~\cite{caba2015activitynet}. VideoChat2~\cite{li2023mvbench} combines multiple existing image and video datasets to develop a 1.9M joint image-video instruction tuning dataset. In our experiments, we use VideoInstruct100K and a subset of the dataset from VideoChat2. Additionally, addressing the limitations of the  VideoInstruct100K dataset~\cite{Maaz2023VideoChatGPT}, we develop~\ourdata~through a novel semi-automatic annotation pipeline, which provides dense video captions along with 112K QA pairs targeting reasoning, spatial and temporal understanding, which further improves model's understanding of video content (see Fig.~\ref{fig:data_block_diagram}).

\textbf{Video Conversation Benchmarks:} Video-ChatGPT~\cite{Maaz2023VideoChatGPT} introduces VCGBench and Zero-shot QA benchmarks, where VCGBench includes 500 videos with 3000 QA pairs, evaluated using GPT-3.5 across various metrics. Despite its comprehensive evaluation, it only contains videos from the ActivityNet dataset. The Zero-shot evaluation covers MSVD-QA~\cite{msvd}, MSR-VTT-QA~\cite{msvd}, TGIF-QA~\cite{TGIF}, and ActivityNet-QA~\cite{caba2015activitynet}. 
MVBench~\cite{li2023mvbench} consists of 4K QA pairs evaluating 20 temporal tasks, though it mostly includes short videos averaging 5-40 seconds. Considering the limitation of existing benchmarks, which often lack focus on generalization and diversity, we propose \ourbench, featuring 4,354 QA pairs from 877 videos across 18 domains (see Fig.~\ref{fig:benchmark_block_diagram}).

\section{Method}
\label{method}
\begin{figure}[!t]
  \centering
    \includegraphics[width=0.99\linewidth]{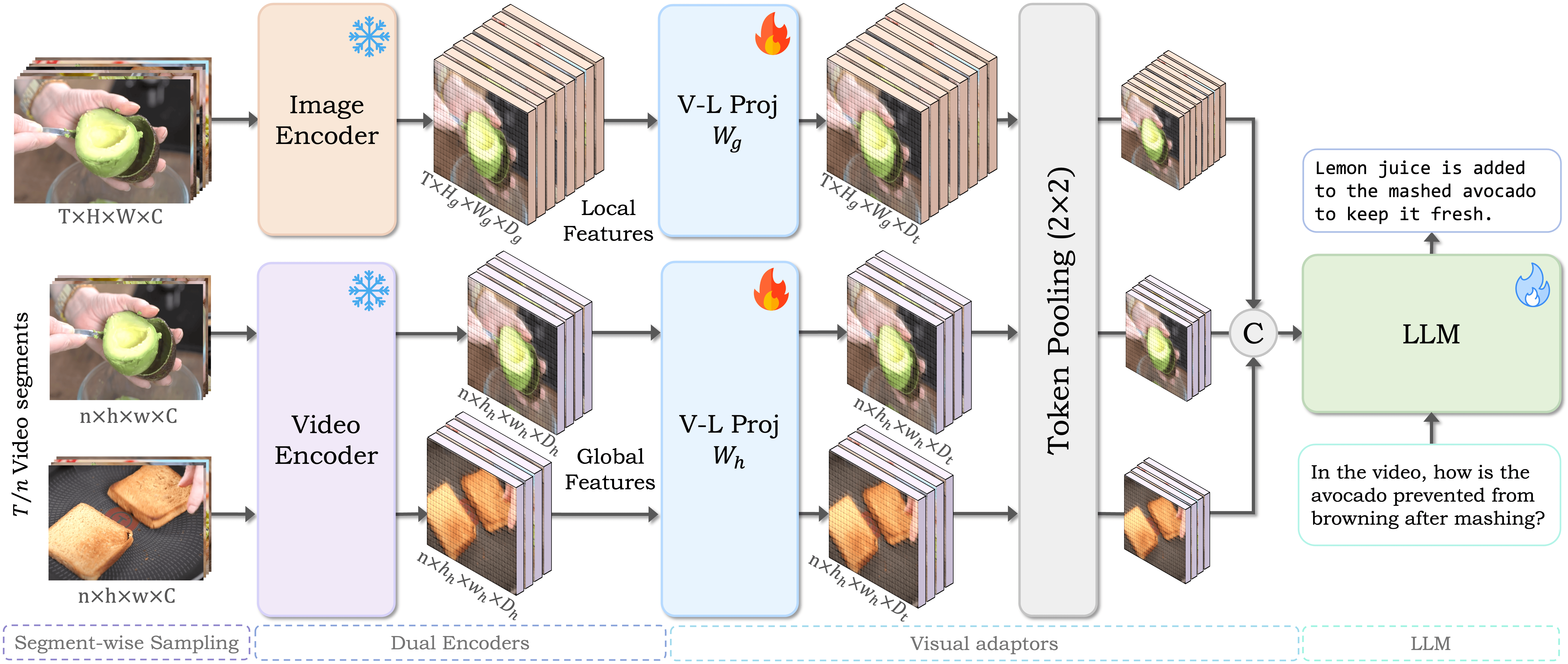}\vspace{-0.8em}
    \caption{\small \textbf{Overview of \ourmodelbold}.~\ourmodel is a large multimodal model for video understanding. It uses a dual-encoder design that combines the complementary strengths of an image encoder and a video encoder. The image encoder captures detailed spatial features, while the video encoder captures temporal dynamics across multiple frames. To retain fine-grained temporal details while ensuring efficiency, we use segment-wise frame sampling instead of random sparse sampling. Both sets of features are then projected into a unified space through Vision-Language (V-L) projection layers and the resulting tokens are pooled and concatenated before being processed by a Large Language Model to generate comprehensive responses to video-based questions. Symbols 
     \includegraphics[height=1em]{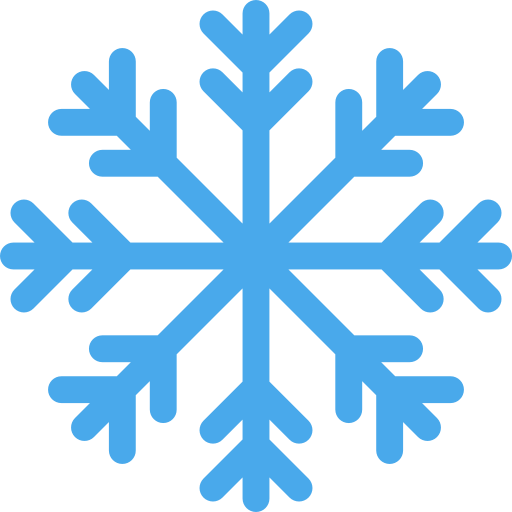} indicates frozen components, \includegraphics[height=1em]{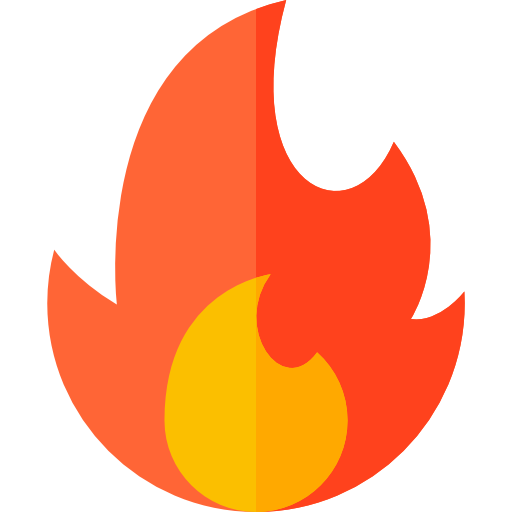} indicates trainable components, and the 
     \includegraphics[height=1em]{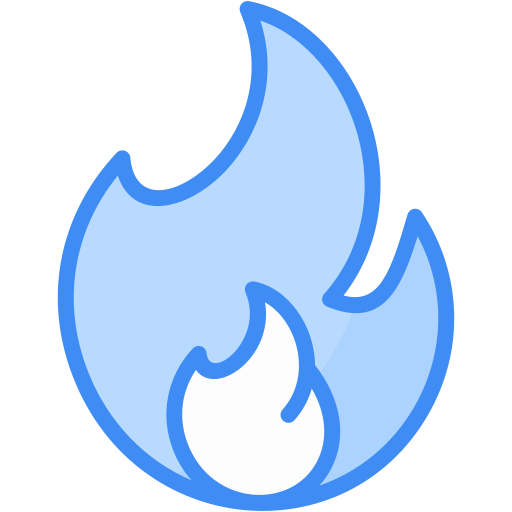} indicates LoRA-training.}
    \label{fig:arch_block_diagram}
\vspace{-1em}
\end{figure}

For effective video understanding, combining detailed spatial information with explicit temporal context is crucial. To achieve this, we propose \ourmodel, which features a dual encoder design that leverages the complementary strengths of an image encoder and a video encoder. 

\noindent 
\textbf{Overall Architecture}:
The overall architecture consists of \rom{1} segment-wise sampling, \rom{2} dual visual encoder, \rom{3} vision-language adapters that project vision features to the language domain and \rom{4} a large language model. Frames selected through a segment-wise sampling strategy are encoded through a dual encoder consisting of an image and a video encoder. Both sets of features are projected to language space using vision-language (V-L) adapters, and the resulting tokens are pooled through adaptive token pooling and concatenated before being fed to the LLM (see Fig.~\ref{fig:arch_block_diagram}).

\noindent
\textbf{Segment-wise Sampling:} To extract fine-grained temporal cues, we use a segment-wise frame sampling strategy. Given an input video $\mathbf{V} \in \mathbb{R}^{T \times H \times W \times C}$, we divide it into $K$ segments, where each segment consists of $n = \frac{T}{K}$ frames. Thus, the video can be represented as $\mathbf{V} = [\mathbf{V}_k]_{k=1}^K$. Each segment $\mathbf{V}_k \in \mathbb{R}^{n \times H \times W \times C}$ can be described as a sequence of frames, $\mathbf{X}_i$, where $\mathbf{V}_k = [\mathbf{X}_{i, j}]_{j=1}^{n}$. The video segments are downsampled to a lower resolution of $n \times h \times w \times c$ for video encoding.

Compared to a uniform sampling, segment-wise sampling better aligns with our dual encoder design. Video encoders often face computational constraints, limiting them to processing only sparse frames. Uniform sampling increases the self-attention computation complexity as it requires attending to features of all frames. Additionally, video encoders are typically trained with sparse frames, and providing more frames can hinder their ability to accurately capture temporal information. In contrast, the segment-wise sampling strategy divides the video into smaller, manageable segments, enabling the video encoder to efficiently capture rich temporal cues within each segment.

\noindent
\textbf{Dual Vision Encoder:} Our design leverages the complementary strengths of an image encoder that captures detailed spatial features and a video encoder that provides explicit temporal context. The image encoder $ g $, processes $T$ frames, $ g(\mathbf{X}) \in \mathbb{R}^{T \times H_g  \times W_g \times D_g} $, producing local features that provide frame-level context. Meanwhile, the video encoder $ h $, operates on low-resolution video segments $ \mathbf{V}_k $, yielding global features that provide segment-wise context, $h(\mathbf{V}_k) \in \mathbb{R}^{n \times h_h  \times w_h \times D_h} $. 

The primary goal of \ourmodel~is to leverage the capabilities of a pre-trained LLM alongside visual modalities from both a pre-trained image encoder and a pre-trained video encoder. Specifically, we utilize the pre-trained CLIP model, ViT-L/14 ($336\times336$)~\cite{clip} as the image encoder, and InternVideo-v2 ($224\times224$)~\cite{wang2024internvideo2} as the video encoder. These models are selected for their robust performance and their ability to complement each other in capturing both spatial and temporal information. Both encoders are pre-trained on large-scale datasets in a multimodal setting using contrastive loss, facilitating their integration within our architecture. 

\noindent
\textbf{Visual Adapter:} The output embeddings from the second last layer of both image and video encoders are passed through separate V-L projection layers, $W_g$ and $W_h$, respectively. These Multi-Layer perceptrons (MLPs) project the visual features into the language space. The projection layers are trainable, while the visual encoders remain frozen, preserving the rich, pre-trained representations. The projected embeddings are reshaped back into their grid forms and subjected to a $2\times2$ adaptive token pooling, which operates on the spatial dimensions of the local and global features. This pooling reduces the token length by a factor of $4$, thereby allowing to fit in larger visual context within the same LLM context window.
The pooled embeddings from the local features form $ \mathbf{E}^{img} \in \mathbb{R}^{T \times h_g  \times w_g \times D_t} $, while the pooled embeddings from the global features of each segment form $ \mathbf{E}^{vid} \in \mathbb{R}^{n \times h_h  \times w_h \times D_t} $. 

\noindent
\textbf{Large Language Model:} We obtain the final representation by concatenating the embeddings $ \mathbf{E}^{img} $ with $K$ segment-wise embeddings $ \mathbf{E}^{vid} $, such that we have detailed spatial representation across all segments followed by their global temporal context. We then concatenate the text embeddings $ \mathbf{E}^{text} \in \mathbb{R}^{L \times D_t} $ of the user text query with the visual embeddings,
\begin{equation}
    \mathbf{E} = [\mathbf{E}^{img}, \mathbf{E}_1^{vid},  \ldots, \mathbf{E}_K^{vid},  \mathbf{E}^{text}].
\end{equation}
This integration ensures that the LLM receives a sequence of embeddings that include detailed spatial features from the image encoder and comprehensive temporal context from the video encoder, allowing for robust video understanding. The LLM is fine-tuned using LoRA~\cite{hu2021lora} in an auto-regressive manner with a next-token prediction loss. Refer to Fig.~\ref{fig:arch_block_diagram} for detailed illustration.

\section{Dataset}
\label{dataset}
\begin{figure}[!t]
  \centering
    \includegraphics[width=0.99\linewidth]{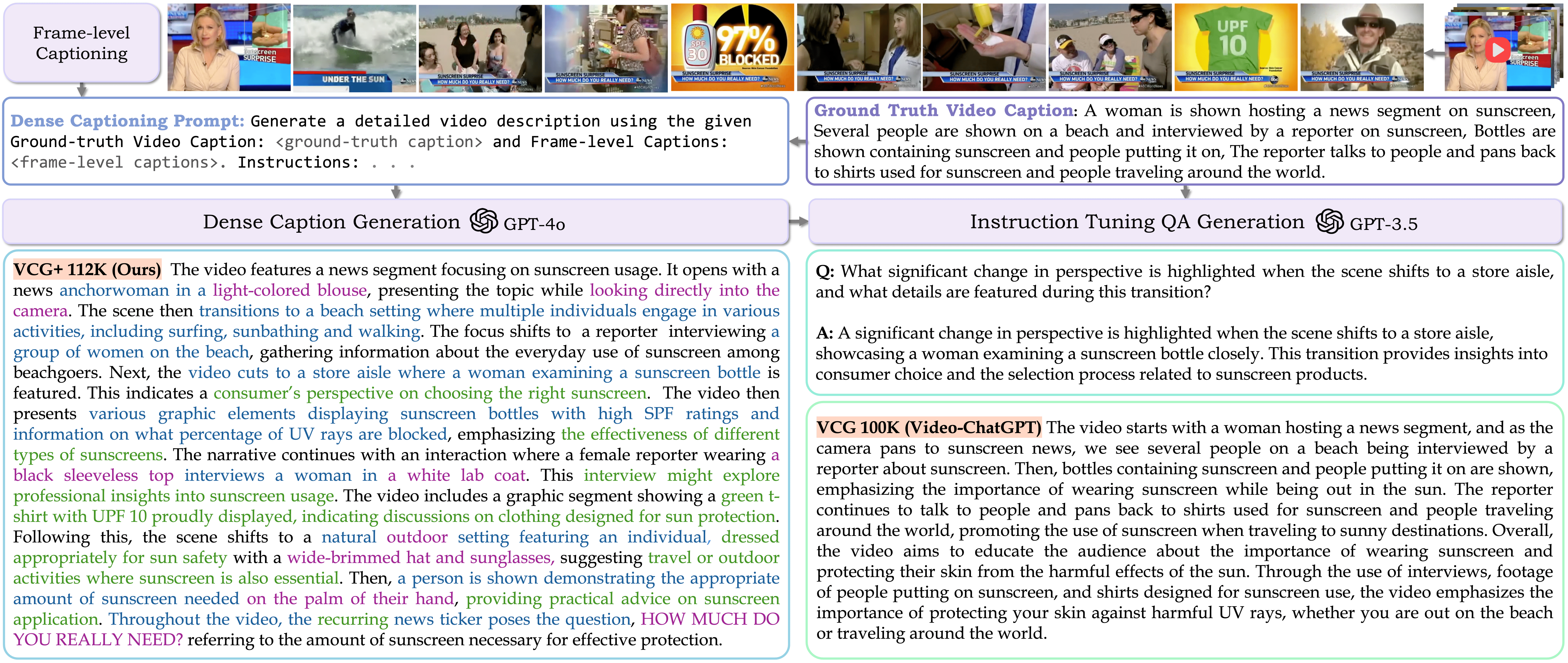}\vspace{-0.8em}
    \caption{\small \textbf{Illustration of the semi-automatic annotation process in \ourdatabold}. 
    The figure shows how we use ground-truth video captions and frame-level descriptions to generate a detailed video description. GPT-4 is used to remove irrelevant and conflicting noisy information in the frame-level descriptions to produce a high-quality video description. The semi-automatic annotation process integrates 
    \textcolor{pink}{spatial}, \textcolor{blue}{temporal and event}, and \textcolor{green}{reasoning} details into the brief information we start with. This dense video description is then used to generate instruction-tuning QA pairs using GPT-3.5. We provide detailed prompts used in both stages in Appendix~\ref{sup:gpt_prompts} (see Figs.~\ref{fig:prompt_1_data} and~\ref{fig:prompt_2_data}). We also compare the video description in the VideoInstruct100K~\cite{Maaz2023VideoChatGPT} dataset to show the improvement in quality achieved by our new annotation pipeline.}
    \label{fig:data_block_diagram}
\vspace{-1em}
\end{figure}

Video-ChatGPT~\cite{Maaz2023VideoChatGPT} introduces the VideoInstruct100K dataset, which employs a semi-automatic annotation pipeline to generate 75K instruction-tuning QA pairs. To address the limitations of this annotation process, we present \ourdata~dataset developed through an improved annotation pipeline. Our approach improves the accuracy and quality of instruction tuning pairs by improving keyframe extraction, leveraging SoTA large multimodal models (LMMs) for detailed descriptions, and refining the instruction generation strategy.

\noindent \textbf{Keyframe Extraction}:
VideoInstruct100K uses a fixed number of video keyframes, regardless of video length or dynamics, to generate frame-level dense captions. This often results in both insufficient and redundant information. We address this by first extracting scenes from videos~\cite{PySceneDetect}, and then selecting one keyframe/scene. Consequently, we obtain 
detailed information for videos with rich content and reduce redundancy for videos with less content. It provides better visual context by extracting more stable keyframes, thus offering a more accurate video representation.

\noindent \textbf{Frame-Level Descriptions}:
After extracting keyframes, we use a SoTA image LMM, LLaVA-v1.6~\cite{liu2024llavanext}, to generate dense descriptions for each keyframe. These descriptions encompass comprehensive visual details, including spatial attributes, scene context, and object characteristics, which are often absent in concise ground truth captions. While ground truth captions are precise, they lack the granularity to capture intricate visual and spatial information. To address this, we augment them captions with detailed but noisy information from the frame-level descriptions, thus enhancing the quality and accuracy of the subsequent video descriptions.

\noindent \textbf{Detailed Video Descriptions}:
VideoInstruct100K~\cite{Maaz2023VideoChatGPT} prompts GPT-3.5 directly with frame-level descriptions and concise ground truth captions to generate QA pairs, imposing a significant cognitive load on the model to verify frame-level descriptions with the ground truth. We improve this process by first creating a coherent and detailed video description. We prompt GPT-4 to integrate the detailed frame-level descriptions with the ground truth captions by comparing information and removing any inconsistencies. The resulting detailed descriptions include a timeline of events, actions, object attributes, and scene settings, providing a thorough representation of the video content. This structured input 
simplifies the task for LLM, thereby enhancing the generated QA pairs quality.

\noindent \textbf{Improved Instruction Tuning Data}: 
Using the ground truth captions and detailed video descriptions, we generate two types of high-quality QA pairs using GPT-3.5: descriptive and concise. For \textbf{descriptive} instruction pairs, we focus on three categories: \rom{1} \textit{dense captioning}, which provides descriptions of the video covering the entire sequence of events and visual details; \rom{2} \textit{detailed temporal information}, which addresses the sequence of events and their dependency to learn temporal relationships; and \rom{3} \textit{generic question answering}, which involves in-depth questions about different actions, their consequences, and other detailed aspects of the video. For \textbf{concise} instruction pairs, we target \rom{1} \textit{spatial reasoning}, focusing on understanding and describing spatial details such as scene settings, number of objects, attire, and locations; \rom{2} \textit{reasoning} of events, covering the causal relationships between events; and \rom{3} \textit{short temporal questions}, addressing specific moments or sequences, such as what happened at the beginning or end.

\section{Proposed Benchmark}
\label{benchmarks}
\begin{figure}[!t]
  \centering
    \includegraphics[width=0.99\linewidth]{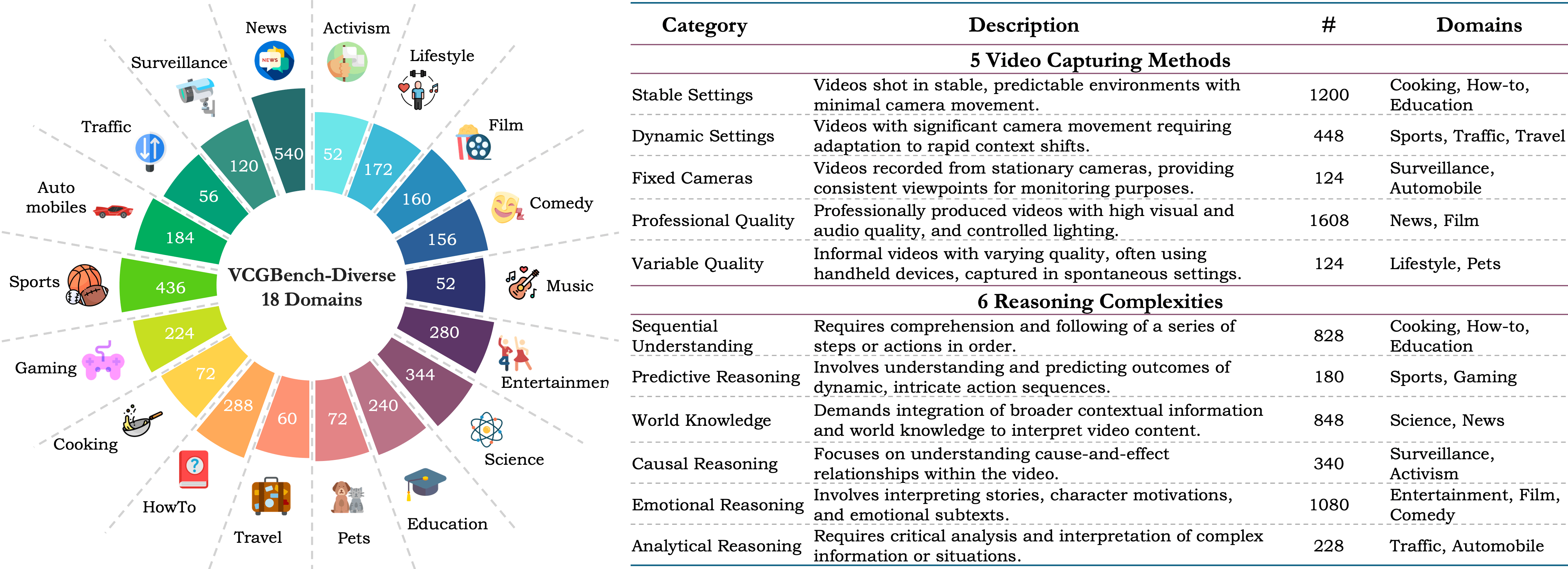}\vspace{-0.8em}
    \caption{\small \textbf{Illustration of \ourbenchbold video conversational benchmark}.  \ourbench comprehensive benchmark is designed to evaluate video LMMs across 18 broad video categories. With 4,354 QA pairs, \ourbench~tests generalization on dense video captioning, spatial and temporal understanding, and complex reasoning. It covers five video-capturing methods, ensuring diversity and robust generalization and six reasoning complexities, assessing various analytical and comprehension skills.}
    \label{fig:benchmark_block_diagram}
\vspace{-1em}
\end{figure}

Recognizing the limited diversity in existing video conversation benchmarks, we introduce \ourbench to comprehensively evaluate the generalization ability of video LMMs. While VCG-Bench~\cite{Maaz2023VideoChatGPT} provides an extensive evaluation protocol, it is limited to videos from the ActivityNet200~\cite{caba2015activitynet} dataset. Our benchmark comprises a total of 877 videos, 18 broad video categories and 4,354 QA pairs, ensuring a robust evaluation framework. The detailed breakdown of \ourbench is illustrated in Fig.~\ref{fig:benchmark_block_diagram}, showcasing the distribution of videos across content domains, video capturing methods, and reasoning complexities.

We collect videos from \textit{18 distinct domains}, including lifestyle, how-to, science and technology, news, travel, entertainment, film, sports, comedy, activism, gaming, education, surveillance, pets, cooking, music, automobile, and traffic (see Fig.~\ref{fig:benchmark_block_diagram}). These categories encompass a broad spectrum of real-world scenarios, ensuring that models are evaluated on a diverse set of challenges. In addition to content diversity, \ourbench includes a variety of \textit{video capture methods}, which ensures a comprehensive assessment of robustness to different filming techniques, camera movements, quality levels and lighting. The benchmark covers \textit{five} video capture methods including static and controlled settings, dynamic and unpredictable settings, fixed camera perspectives, professional and high-quality videos, and uncontrolled and variable quality. Further, the benchmark evaluates models across \textit{six reasoning complexities}, including sequential understanding, complex action and predictive reasoning, contextual and world knowledge reasoning, causal reasoning, narrative and emotional reasoning, and analytical and critical reasoning, which is crucial for understanding diverse video content. 

The videos in \ourbench are sourced from HDVILA~\cite{xue2022hdvila}, MPII~\cite{andriluka14cvpr}, YouCook2~\cite{zhou2018towards}, UCF Crime~\cite{ucfcrime}, and STUD Traffic~\cite{xu2021sutd}. The video durations range from 29 sec to 471 sec, with an average of 217 sec. Human annotators are tasked with writing detailed descriptions based on their understanding of both audio and visual elements of the videos. This comprehensive annotation process involves a set of annotators who are provided with an initial set of ten videos each. These annotations undergo a meta-review stage where feedback is provided, and necessary corrections are made to meet the required standards. Following this, annotators receive additional batches, with random samples being selected for quality checks by the meta-reviewer. The final human annotations are utilized to generate QA pairs using GPT-3.5, based on prompts detailed in Fig.~\ref{fig:vcgbench_diverse_dataset_generation}. 

Following VCG-Bench~\cite{Maaz2023VideoChatGPT}, the evaluation is computed over five different aspects: \rom{1} correctness of information \rom{2} detail orientation \rom{3} contextual understanding \rom{4} temporal understanding and \rom{5} consistency. Additionally, \ourbench provides a breakdown of performance across three key aspects: \rom{1} dense video captioning, which assesses the ability to generate detailed and accurate descriptions of the video content, \rom{2} spatial understanding, which evaluates the capability to understand and describe the spatial relationships and settings within the video, and \rom{3} reasoning, which tests the adeptness in inferring and explaining causal relationships and actions within the video.

\section{Experiments}
\label{sec:experiments}

We perform quantitative evaluation of \ourmodel~on four standard benchmarks: i) \texttt{VCGBench}~\cite{Maaz2023VideoChatGPT}, ii) \ourbench, iii) \texttt{MVBench}~\cite{li2023mvbench} and iv) Zero-shot QA.

\noindent
\textbf{Implementation Details:} We use CLIP-L/14~\cite{clip} as our image encoder, InternVideo-v2~\cite{wang2024internvideo2} stage-2 1B model as our video encoder in conjunction with Phi-3-Mini-3.8B~\cite{phi3mini4k} based LLM with 4K context window in our experiments. The image encoder operates at $336\times336$, while the video encoder operates at $224\times224$ resolution. Our training consists of two pretraining stages and one instruction-tuning stage. In the pretraining stage, we train with only the image encoder and only the video encoder on the CC-595K dataset~\cite{liu2023improved}, with only the visual adapters being learned while the rest of the model is kept frozen. During the instruction-tuning stage, we use LoRA~\cite{hu2022lora} with $r=64$ for LLM, while visual adapters are fully trained and vision encoders are kept frozen. The LR is set to $1e^{-3}$ during pretraining and $2e^{-4}$ during instruction tuning.

For experiments on \texttt{VCGBench}, \ourbench~and Zero-shot QA, we sample 16 frames from videos, while for MVBench which consists of relatively shorter videos, we sample 8 frames. We keep the same sampling strategy during inference. For \texttt{VCGBench} and~\ourbench, the model is trained on VideoInstruct100K~\cite{Maaz2023VideoChatGPT}, \ourdata~, conversation and caption data from VideoChat~\cite{2023videochat} and VQA dataset from WebVid~\cite{bain2021frozen}, that combines to approximately 260K single turn conversations. For \texttt{MVBench}, the model is trained on Kinetics-710~\cite{kay2017kinetics}, Something-Something-v2~\cite{goyal2017something}, conversations from VideoChat~\cite{2023videochat}, CLEVRER~\cite{yi2019clevrer}, VQA dataset from WebVid~\cite{bain2021frozen} and NExT-QA~\cite{xiao2021next} datasets, which combines to approximately 330K single turn conversations. We run all trainings for one epoch. Following previous approaches~\cite{Maaz2023VideoChatGPT, jin2023chatunivi, st-llm}, we employ GPT-3.5-Turbo-0613 for \texttt{VCGBench} and Zero-shot QA evaluation. However, for our proposed \ourbench, we employ the latest GPT-3.5-Turbo-0125 for evaluation.

\begin{wraptable}{r}{8.3cm}
\vspace{-1em}
\resizebox{0.6\textwidth}{!}{
\footnotesize
{
\begin{tabular}{lcccccc}
\toprule
\textbf{Method} & \textbf{CI} & \textbf{DO} & \textbf{CU} & \textbf{TU} & \textbf{CO} & \textbf{Avg.} \\ \midrule

Video-ChatGPT~\cite{Maaz2023VideoChatGPT}            & 2.40             & 2.52            & 2.62             & 1.98              & 2.37             & 2.38                \\
BT-Adapter~\cite{bt_adapter}               & 2.68             & 2.69            & 3.27             & 2.34              & 2.46             & 2.69                \\
VTimeLLM~\cite{huang2023vtimellm}                 & 2.78             & \underline{3.10}            & 3.40             & 2.49              & 2.47             & 2.85                \\
Chat-UniVi~\cite{jin2023chatunivi}               & 2.89             & 2.91            & 3.46             & \textbf{2.89}              & 2.81             & 2.99                \\
LLAMA-VID~\cite{llamavid}                & 2.96             & 3.00            & \underline{3.53}             & 2.46              & 2.51             & 2.89                \\
Video-LLaVA~\cite{video-llava}                & 2.84             & 2.86            & 3.44             & 2.46              & 2.57             & 2.81                \\
VideoChat2~\cite{li2023mvbench}               & 3.02             & 2.88            & 3.51             & 2.66              & 2.81             & 2.98                \\ 
IG-VLM~\cite{kim2024image}                   & \underline{3.11}             & 2.78            & 3.51             & 2.44              & \underline{3.29}             & \underline{3.03}                \\

\rowcolor{violet!10} \ourmodel~(ours) & \textbf{3.27}            & \textbf{3.18}            & \textbf{3.74}             & \underline{2.83}              & \textbf{3.39}             & \textbf{3.28}                \\
\bottomrule
\end{tabular}
}
}
\caption{\small \textbf{Performance of \ourmodelbold~on VCGBench~\cite{Maaz2023VideoChatGPT}.} All models use 16 frames except Video-ChatGPT and Chat-UniVi which use 100 and 64 frames respectively. 
}
\label{tab:vcgbench}
\vspace{-1em}
\end{wraptable}

\noindent
\textbf{VCGBench:} The benchmark consists of approximately 3000 QA pairs generated using 500 human-annotated videos from ActivityNet~\cite{caba2015activitynet}. The benchmark evaluates the responses on five different aspects: i) Correctness of Information (CI), which assesses the correctness of the response to ensure it aligns with the video contents, ii) Detail Orientation (DO), which evaluates the depth of the response, iii) Contextual Understanding (CU), which assesses if the response aligns with the overall context of the video, iv) Temporal Understanding (TU), which assesses the model's ability to identify temporal sequences accurately, and v) Consistency (CO), which evaluates the consistency in the model response to similar questions. Table~\ref{tab:vcgbench} compares our model with previous SoTA approaches. \ourmodel~achieves an average score of 3.28 surpassing previous best method by a margin of 0.25 (5\%).


\textbf{VCGBench-Diverse:} We provide a quantitative comparison of \ourmodel~against previous SoTA approaches on \ourbench, which contains 4,354 QA pairs from 877 videos. Following \cite{Maaz2023VideoChatGPT}, we evaluate the Correctness of Information (CI), Detail Orientation (DO), Contextual Understanding (CU), Temporal Understanding (TU), and Consistency (CO). Additionally, we provide results for dense captioning, spatial understanding, and visual reasoning abilities. The results are presented in Table~\ref{tab:vcgbench_diverse}. \ourmodel~achieves an average score of 2.47 surpassing all previous methods. Further, \ourmodel~achieves a score of 1.38, 2.80, and 3.63 on dense captioning, spatial understanding, and visual reasoning, respectively. Notably, \ourmodel achieves improvements in spatial and temporal understanding, surpassing previous best models by 0.37 (7.4\%) and 0.23 (4.6\%), respectively. This is attributed to the dual encoder architecture, where the high-resolution image encoder enhances spatial understanding and the video encoder improves temporal accuracy.

\begin{table*}[t!]
\centering
\resizebox{0.99\textwidth}{!}{
\begin{tabular}{lcccccc|ccc}
\toprule
\textbf{Method} & \textbf{CI} & \textbf{DO} & \textbf{CU} & \textbf{TU} & \textbf{CO} & \textbf{Avg.} & \textbf{Caption} & \textbf{Spatial} & \textbf{Reasoning} \\ \midrule
Video-ChatGPT~(ACL 2024)~\cite{Maaz2023VideoChatGPT}            & 2.07             & 2.42            & 2.46             & 1.39              & 2.06             & 2.08                & 0.89             & 2.25             & 3.60       \\
BT-Adapter~(CVPR 2024)~\cite{bt_adapter}               & 2.20             & \underline{2.62}            & 2.59             & 1.29              & 2.27             & 2.19                & 1.03             & 2.35             & \underline{3.62}       \\
VTimeLLM~(CVPR 2024)~\cite{huang2023vtimellm}               & 2.16             & 2.41            & 2.48             & 1.46              & 2.35             & 2.17                & 1.13             & 2.29               & 3.45        \\
Chat-UniVi~(CVPR 2024)~\cite{jin2023chatunivi}               & \underline{2.29}             & 2.56            & \underline{2.66}             & 1.56              & \underline{2.36}             & \underline{2.29}                & \underline{1.33}             & 2.36             & 3.59        \\

VideoChat2~(CVPR 2024)~\cite{li2023mvbench}               & 2.13             & 2.42            & 2.51             & \underline{1.66}              & 2.27             & 2.20                & 1.26             & \underline{2.43}             & 3.13      \\ 

\rowcolor{violet!10} \ourmodel~(ours)        & \textbf{2.46}             & \textbf{2.73}            & \textbf{2.81}             & \textbf{1.78}              & \textbf{2.59}             & \textbf{2.47}                & \textbf{1.38}             & \textbf{2.80}             & \textbf{3.63}        \\

\bottomrule
\end{tabular}
}
\caption{\small \textbf{Performance of \ourmodelbold~on VCGBench-Diverse.} All models use 16 frames except Video-ChatGPT and Chat-UniVi, which use 100 and 64 frames, respectively. The good performance of our model on ~\ourbench shows its generalization to diverse scenarios.}
\label{tab:vcgbench_diverse}
\end{table*}

\begin{table*}[h!]
    \centering
    \setlength\tabcolsep{2pt}
    \resizebox{1.0\textwidth}{!}{
        \begin{tabular}{lccccccccccccccccccccc}
        \toprule
        \textbf{Model} & \textbf{AS} & \textbf{AP} & \textbf{AA} & \textbf{FA} & \textbf{UA} & \textbf{OE} & \textbf{OI} & \textbf{OS} & \textbf{MD} & \textbf{AL} & \textbf{ST} & \textbf{AC} & \textbf{MC} & \textbf{MA} & \textbf{SC} & \textbf{FP} & \textbf{CO} & \textbf{EN} & \textbf{ER} & \textbf{CI} & \cellcolor{gray!20}\textbf{Avg.} \\
        \midrule
        Random & 25.0 & 25.0 & 33.3 & 25.0 & 25.0 & 33.3 & 25.0 & 33.3 & 25.0 & 25.0 & 25.0 & 33.3 & 25.0 & 33.3 & 33.3 & 25.0 & 33.3 & 25.0 & 20.0 & 30.9 & \cellcolor{gray!20}{27.3} \\
        GPT-4V~\cite{2023GPT4VisionSC} & 55.5 & 63.5 & 72.0 & 46.5 & 73.5 & 18.5 & 59.0 & 29.5 & 12.0 & 40.5 & 83.5 & 39.0 & 12.0 & 22.5 & 45.0 & 47.5 & 52.0 & 31.0 & 59.0 & 11.0 & \cellcolor{gray!20}{43.5} \\
        \midrule
        Otter-V~\cite{li2023otter} & 23.0 & 23.0 & 27.5 & 27.0 & 29.5 & 53.0 & 28.0 & 33.0 & 24.5 & 23.5 & 27.5 & 26.0 & 28.5 & 18.0 & 38.5 & 22.0 & 22.0 & 23.5 & 19.0 & 19.5 & \cellcolor{gray!20}{26.8} \\
        mPLUG-Owl-V~\cite{ye2023mplug} & 22.0 & 28.0 & 34.0 & 29.0 & 29.0 & 40.5 & 27.0 & 31.5 & \underline{27.0} & 23.0 & 29.0 & 31.5 & 27.0 & 40.0 & 44.0 & 24.0 & 31.0 & 26.0 & 20.5 & 29.5 & \cellcolor{gray!20}{29.7} \\
        Video-ChatGPT~\cite{Maaz2023VideoChatGPT} & 23.5 & 26.0 & 62.0 & 22.5 & 26.5 & 54.0 & 28.0 & \underline{40.0} & 23.0 & 20.0 & 31.0 & 30.5 & 25.5 & 39.5 & \textbf{48.5} & 29.0 & 33.0 & \underline{29.5} & 26.0 & 35.5 & \cellcolor{gray!20}{32.7} \\
        VideoLLaMA~\cite{damonlpsg2023videollama} & 27.5 & 25.5 & 51.0 & 29.0 & 39.0 & 48.0 & 40.5 & 38.0 & 22.5 & 22.5 & 43.0 & 34.0 & 22.5 & 32.5 & 45.5 & 32.5 & 40.0 & 30.0 & 21.0 & 37.0 & \cellcolor{gray!20}{34.1} \\
        VideoChat~\cite{2023videochat} & 33.5 & 26.5 & 56.0 & 33.5 & 40.5 & 53.0 & 40.5 & 30.0 & 25.5 & \underline{27.0} & 48.5 & 35.0 & 20.5 & 42.5 & \underline{46.0} & 26.5 & \underline{41.0} & 23.5 & 23.5 & 36.0 & \cellcolor{gray!20}{35.5} \\
        VideoChat2~\cite{li2023mvbench} & \underline{66.0} & \underline{47.5} & \textbf{83.5} & \textbf{49.5} & \underline{60.0} & \underline{58.0} & \underline{71.5} & \textbf{42.5} & 23.0 & 23.0 & \underline{88.5} & \underline{39.0} & \underline{42.0} & \underline{58.5} & 44.0 & \underline{}{49.0} & 36.5 & \textbf{35.0} & \underline{40.5} & \textbf{65.5} & \cellcolor{gray!20}{\underline{51.1}} \\
        \rowcolor{violet!10} \ourmodel~(ours) & \textbf{69.0} & \textbf{60.0} & \underline{83.0} & \underline{48.5} & \textbf{66.5} & \textbf{85.5} & \textbf{75.5} & 36.0 & \textbf{44.0} & \textbf{34.0} & \textbf{89.5} & \textbf{39.5} & \textbf{71.0} & \textbf{90.5} & 45.0 & \textbf{53.0} & \textbf{50.0} & \underline{29.5} & \textbf{44.0} & \underline{60.0} & \cellcolor{gray!20}{\textbf{58.7}} \\
        
        \bottomrule
        \end{tabular}
    }
    \caption{\small \textbf{Performance of \ourmodelbold~on MVBench.} Following~\cite{li2023mvbench}, we evaluate on 20 tasks including \underline{AS}: Action Sequence, \underline{AP}: Action Prediction, \underline{AA}: Action Antonym, \underline{FA}: Fine-grained Action, \underline{UA}: Unexpected Action, \underline{OE}: Object Existence, \underline{OI}: Object Interaction, \underline{OS}: Object Shuffle, \underline{MD}: Moving Direction, \underline{AL}: Action Localization, \underline{ST}: Scene Transition, \underline{AC}: Action Count, \underline{MC}: Moving Count, \underline{MA}: Moving Attribute, \underline{SC}: State Change, \underline{FP}: Fine-grained Pose, \underline{CO}: Character Order, \underline{EN}: Egocentric Navigation, \underline{ER}: Episodic Reasoning and \underline{CI}: Counterfactual Inference.}
\label{tab:mvbench}
\end{table*}

\textbf{MVBench:} We evaluate \ourmodel~on MVBench~\cite{li2023mvbench}, which provides 4,000 QA pairs from 11 video datasets covering a broad spectrum of scenes, ranging from first-person to third-person and from indoor to outdoor environments. The tasks are categorized into 20 fine-grained temporal understanding tasks. The results presented in Table~\ref{tab:mvbench} compare \ourmodel~with previous methods, indicating an overall improvement of 7.6\% compared to the previous best, VideoChat2. Specifically, \ourmodel achieves SoTA results in 14 out of 20 tasks and comes second in 4 out of 20 tasks, obtaining an average score of 58.7\% across the 20 tasks. Additionally, \ourmodel~shows significant improvements in the Action Prediction (+12.5\%), Object Existence (OE) (+27.5\%), Moving Direction (MD) (+17\%), Moving Count (MC) (+29\%) and Moving Attributes (MA) (+32\%) indicating the rich spatial information and temporal context achieved by our model.

\textbf{Zero-shot Question-Answering:} We provide a quantitative comparison of our method on the zero-shot QA task across four open-ended QA datasets, including MSVD-QA~\cite{msvd}, MSRVTT-QA~\cite{msvd}, TGIF-QA~\cite{TGIF}, and ActivityNet-QA~\cite{caba2015activitynet}. Results presented in Table~\ref{tab:zeroshot} show \ourmodel~achieves superior performance compared to previous methods, indicating its ability to adapt effectively to unseen videos and generate accurate contextually relevant responses in challenging settings.

\begin{table*}[b!]
\centering
\setlength{\tabcolsep}{8pt}
\renewcommand{\arraystretch}{1}
\resizebox{0.99\textwidth}{!}{
\begin{tabular}{l c c c c c c c c}
\toprule
\textbf{Model} & \multicolumn{2}{c}{\textbf{MSVD-QA}} & \multicolumn{2}{c}{\textbf{MSRVTT-QA}} & \multicolumn{2}{c}{\textbf{TGIF-QA}} & \multicolumn{2}{c}{\textbf{ActivityNet-QA}} \\
\cmidrule{2-9}
 & \textbf{Accuracy} & \textbf{Score} & \textbf{Accuracy} & \textbf{Score} & \textbf{Accuracy} & \textbf{Score} & \textbf{Accuracy} & \textbf{Score} \\
\midrule
FrozenBiLM~\cite{yang2022frozenbilm} & 32.2 & -- & 16.8 & -- & 41.0 & -- & 24.7 & -- \\
VideoChat~\cite{2023videochat} & 56.3 & 2.8 & 45.0 & 2.5 & 34.4 & 2.3 & 26.5 & 2.2 \\
LLaMA Adapter~\cite{llama_adapter} & 54.9 & 3.1 & 43.8 & 2.7 & - & - & 34.2 & 2.7 \\
Video-LLaMA~\cite{damonlpsg2023videollama} & 51.6 & 2.5 & 29.6 & 1.8 & - & - & 12.4 & 1.1 \\
Video-ChatGPT~\cite{Maaz2023VideoChatGPT} & 64.9 & 3.3 & 49.3 & 2.8 & 51.4 & 3.0 & 35.2 & 2.8 \\
ChatUniVi~\cite{jin2023chatunivi} & 65.0 & 3.6 & 54.6 & 3.1 & 60.3 & 3.4 & 45.8 & 3.2 \\
LLaMA-VID~\cite{llamavid} & 70.0 & \underline{3.7} & 58.9 & 3.3 & -- & -- & 47.5 & \underline{3.3} \\
Video-LLaVA~\cite{video-llava} & \underline{70.7} & \textbf{3.9} & \underline{59.2} & \underline{3.5} & \underline{70.0} & \underline{4.0} & 45.3 & \underline{3.3} \\
VideChat2~\cite{li2023mvbench} & 70.0 & \textbf{3.9} & 54.1 & 3.3 & -- & -- & \underline{49.1} & \underline{3.3} \\
\rowcolor{violet!10} \ourmodel~(ours) & \textbf{72.4} & \textbf{3.9} & \textbf{60.6} & \textbf{3.6} & \textbf{74.6} & \textbf{4.1} & \textbf{50.6} & \textbf{3.6} \\
\bottomrule
\end{tabular}}
\caption{\small \textbf{Performance of \ourmodelbold~on Zero-shot QA.} We evaluate accuracy and score on four commonly used datasets. All the models are evaluated in zero-shot setting where none of the videos were included in the training set. \ourmodel~achieves good results on all datasets.}
\label{tab:zeroshot}
\end{table*}


\begin{table*}[t!]
\centering
\setlength{\tabcolsep}{8pt}
\renewcommand{\arraystretch}{1}
\resizebox{0.99\textwidth}{!}{
\begin{tabular}{lccc|ccc|cc|cc}
\toprule
\multirow{2}{*}{\textbf{}} & \multicolumn{3}{c}{\textbf{Vision Encoder Type}} & \multicolumn{3}{c}{\textbf{Image Pooling}} & \multicolumn{2}{c}{\textbf{Video Pooling}} & \multicolumn{2}{c}{\textbf{VCG+~112K}} \\ 
\cmidrule(lr){2-4} \cmidrule(lr){5-7} \cmidrule(lr){8-9} \cmidrule(lr){10-11}
                  & \textbf{Image} & \textbf{Video} & \textbf{Dual} & \textbf{CNN} & $\boldsymbol{4\times4}$ & $\boldsymbol{2 \times 2}$ & \textbf{Time} & \textbf{Space} & \textbf{\texttimes} & \textbf{\checkmark} \\ 
\midrule
Correctness (CI)       & 3.14     &   3.22     & \cellcolor{violet!10}\textbf{3.27}         & 3.24     &   3.24     & \cellcolor{violet!10}\textbf{3.27}         & 3.21             &\cellcolor{violet!10}\textbf{ 3.27}         & 3.20 & \cellcolor{violet!10}\textbf{3.27} \\
Detail (DO)       & 3.09     &   3.10     & \cellcolor{violet!10}\textbf{3.18}         & 3.13     &   3.18     & \cellcolor{violet!10}\textbf{3.18}         & 3.13             & \cellcolor{violet!10}\textbf{3.18}         & 3.08 & \cellcolor{violet!10}\textbf{3.18} \\
Context (CU)       & 3.68     &   3.70     & \cellcolor{violet!10}\textbf{3.74}         & 3.70     &   3.73     & \cellcolor{violet!10}\textbf{3.74}         & 3.70             & \cellcolor{violet!10}\textbf{3.74}         & 3.66 & \cellcolor{violet!10}\textbf{3.74} \\
Temporal (TU)       & 2.69     &   2.70     & \cellcolor{violet!10}\textbf{2.83}         & 2.74     &   2.73     & \cellcolor{violet!10}\textbf{2.83}        & 2.72             & \cellcolor{violet!10}\textbf{2.83}         & 2.66 & \cellcolor{violet!10}\textbf{2.83} \\
Consistency (CO)       & 3.26     &   3.31     & \cellcolor{violet!10}\textbf{3.39}         & 3.41     &   3.39     & \cellcolor{violet!10}\textbf{3.39}         & 3.36             & \cellcolor{violet!10}\textbf{3.39}         & 3.28 & \cellcolor{violet!10}\textbf{3.39} \\
Average   & 3.17    &  3.20   & \cellcolor{violet!10}\textbf{3.28}   & 3.25    &  3.25   & \cellcolor{violet!10}\textbf{3.28}   & 3.23         & \cellcolor{violet!10}\textbf{3.28}        & 3.17 & \cellcolor{violet!10}\textbf{3.28} \\
\bottomrule
\end{tabular}
}
\caption{\small \textbf{Ablation on Vision Encoder type, Image feature pooling, Video feature pooling, and VCG+ 112K.} We evaluate \ourmodel~training using only an image encoder, video encoder, and both image and video encoders (dual encoder) on VCGBench. Using a dual encoder provides rich semantic and temporal cues.
}
\label{tab:encoder_pooling_and_data}
\vspace{-1em}
\end{table*}

\textbf{Vision Encoder Type:} We ablate our dual visual encoder design in \ourmodel~in on VCGBench with results presented in Table~\ref{tab:encoder_pooling_and_data}. We conduct three experiments: using only the image encoder, only the video encoder, and both encoders. The image encoder alone achieves a score of 3.17, while the video encoder alone achieves a better score of 3.20, indicating the benefits of video-based pretraining. The dual encoder design, combining both spatial and temporal information, achieves the highest score of 3.28, demonstrating enhanced performance in video-conversation tasks.

\textbf{Pooling Strategy:} We ablate different pooling strategies for the image and video encoders in Table~\ref{tab:encoder_pooling_and_data}. The image encoder outputs a $24 \times 24$ feature map from a $336 \times 336$ input. We compare two downsampling methods: a learnable lightweight CNN (LDPv2 from \cite{chu2024mobilevlm}) and a non-learnable adaptive average pooling with a $2 \times 2$ kernel. Results indicate that adaptive pooling performs better than CNN. A $4 \times 4$ adaptive pooling was also tested but showed inferior performance. 

Similarly, we ablate the pooling choice for the video encoder, which takes an input of size $T \times 224 \times 224 \times C$ and outputs a feature map of $T \times 16 \times 16 \times d$. We compare two pooling strategies: time pooling across the temporal dimension to reduce the feature map to $1 \times 16 \times 16 \times d$, and space pooling across the spatial dimension with a $2 \times 2$ kernel. Table~\ref{tab:encoder_pooling_and_data} shows that space pooling effectively preserves temporal information and yields better results.

\begin{wraptable}{r}{7.5cm}
\vspace{-1em}
\resizebox{0.55\textwidth}{!}{
\footnotesize
{
\begin{tabular}{lcccccc}
\toprule
\multirow{2}{*}{\textbf{LLM}} & \multicolumn{5}{c}{\textbf{VCGBench}} & \multirow{2}{*}{\textbf{Avg.}} \\ \cmidrule{2-6}
                                         & \textbf{CI} & \textbf{DO} & \textbf{CU} & \textbf{TU} & \textbf{CO} &  \\ \midrule

Phi3-Mini-3.8B                                & 3.27 & 3.18 & \underline{3.74} & 2.83 & 3.39 & 3.28 \\
Vicuna-7B                                     & 3.22 & 3.14 & 3.69 & 2.65 & \underline{3.46} & 3.23 \\
Vicuna-13B                                    & \textbf{3.30} & \underline{3.20} & \textbf{3.75} & 2.77 & \textbf{3.48} & \textbf{3.30} \\
LLaMA3-8B               & \underline{3.29} & \textbf{3.21} & 3.73 & \textbf{2.86} & 3.38 & \underline{3.29} \\
\bottomrule
\end{tabular}
}
}
\caption{\small \textbf{Ablation on LLM type.} We train and evaluate \ourmodel~with different LLMs, including vicuna~\cite{vicuna2023} and LLaMA3~\cite{llama3}, which further improves accuracy.
}
\label{tab:llm_type}
\vspace{-1em}
\end{wraptable}

\textbf{VCG+ 112K:} To demonstrate the effectiveness of \ourdata, we train \ourmodel with and without it. As shown in Table~\ref{tab:encoder_pooling_and_data}, \ourdata improves performance, particularly in detail orientation (DO) and temporal understanding (TU). This improvement can be attributed to our novel semi-automatic annotation pipeline and the enhanced instruction tuning data, which focuses on generating both detailed and concise instruction pairs. Refer to Fig.~\ref{fig:data_block_diagram} for qualitative visualization of the data.

\textbf{LLM Type:} We train \ourmodel with different LLMs including Vicuna 7B and 13B~\cite{vicuna2023} and LLaMA-3 8B~\cite{llama3} and shows results in Table~\ref{tab:llm_type}. We observe slight improvements in VCGBench scores when training using better LLMs, including Vicuna 13B and LLaMA-3 8B models.

\section{Conclusion}
\label{sec:conclusion}
In this work, we introduce~\ourmodel, a novel video conversation model that leverages the complementary benefits of image and video encoders to achieve enhanced video understanding. \ourmodel~demonstrates better performance across multiple video benchmarks, owing to its dual-encoder design, lightweight visual adapters that map image/video features to a common space and a segment-wise sampling strategy that retains fine-grained temporal information.
We also develop \ourdata, a 112K video-instruction set using a resource-efficient semi-automated annotation pipeline that delivers further gains. Lastly, we propose \ourbench, a diverse benchmark covering 18 video categories, to comprehensively evaluate video LMMs. Despite reported improvements, video LMMs still find challenges in precise action localization, understanding very long videos, and navigating long paths; areas where major improvements can unlock new applications.   

\begin{figure}[!h]
  \centering
    \includegraphics[width=1.0\linewidth]{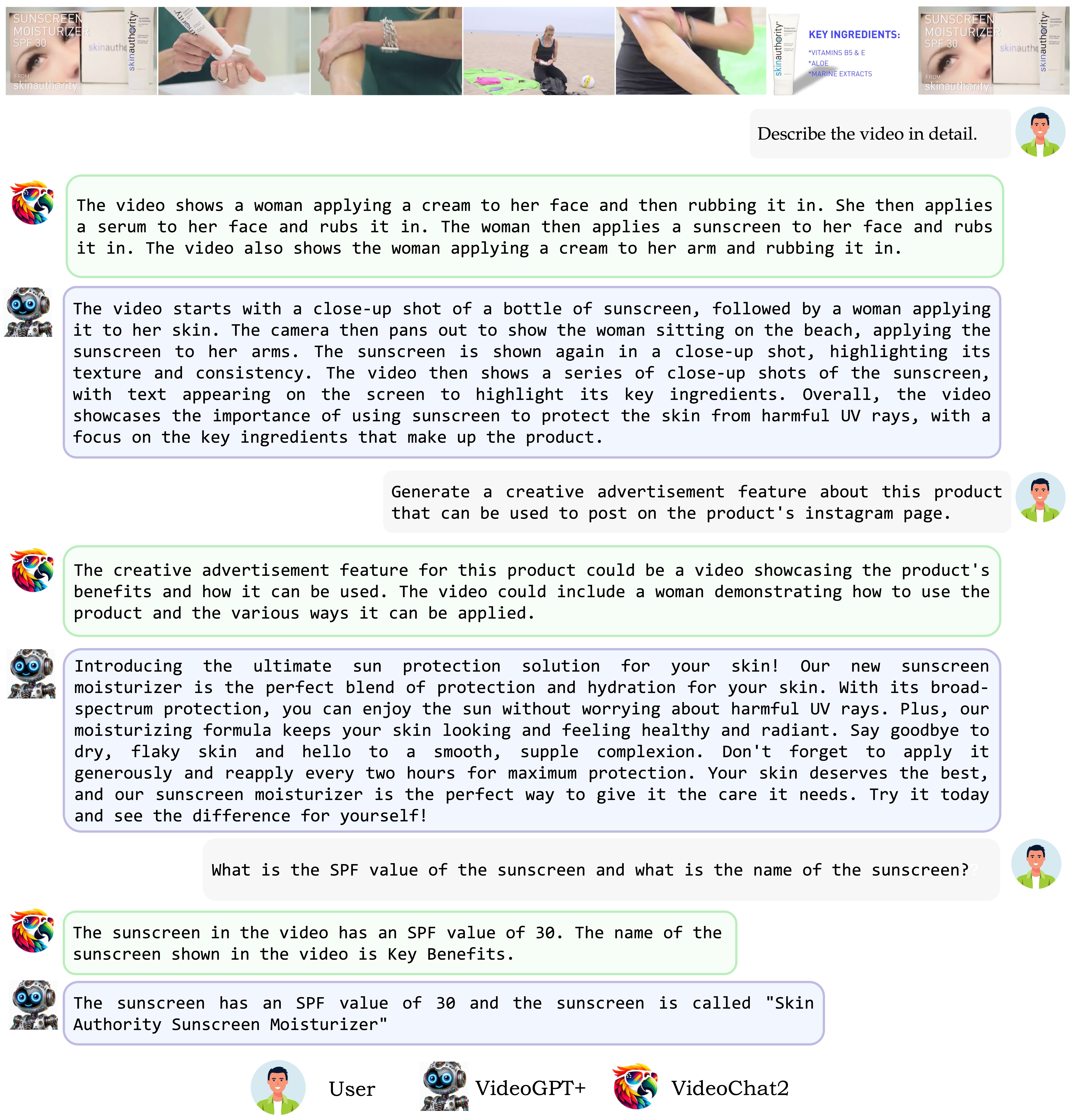}
    \caption{\small \textbf{Qualitative comparison of \ourmodelbold with VideoChat2.} Our \ourmodel demonstrates superior temporal understanding by correctly identifying multiple events in the video, effective reasoning in generating a creative advertisement, and accurate spatial understanding by identifying the SPF value and brand name of the sunscreen.}
    \label{fig:qualitative_1}
\vspace{-1em}
\end{figure}

\begin{figure}[!h]
  \centering
    \includegraphics[width=1.0\linewidth]{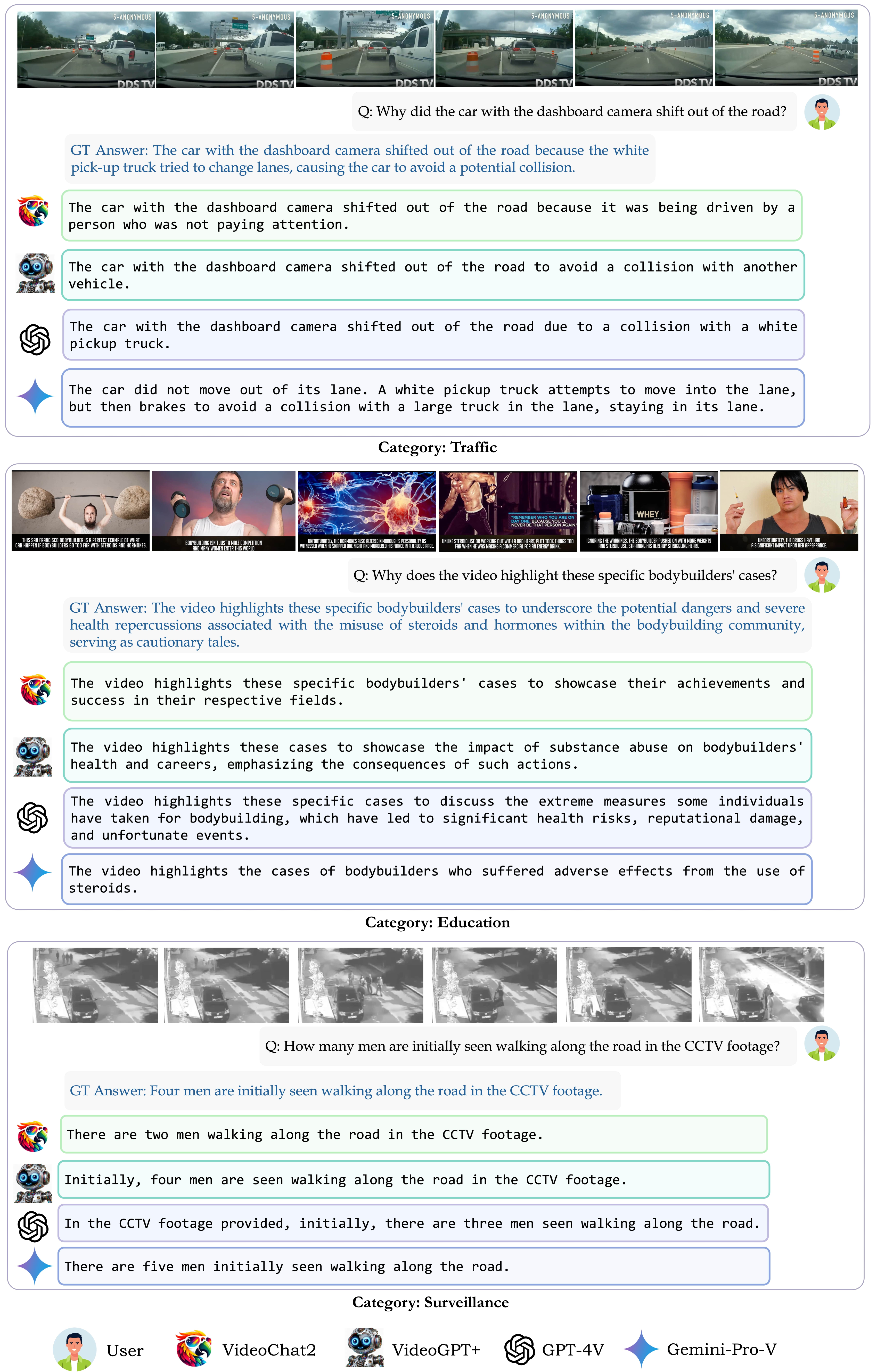}
    \caption{\small \textbf{Qualitative comparison from \ourbench of \ourmodelbold.} We show qualitative comparison of \ourmodel with VideoChat2  and propriety models GPT-4V and Gemini-1.5-Pro-V from three different categories including traffic, education and surveillance from \ourbench.}
    \label{fig:qualitative_3}
\vspace{-1em}
\end{figure}

\begin{figure}[!h]
  \centering
    \includegraphics[width=1.0\linewidth]{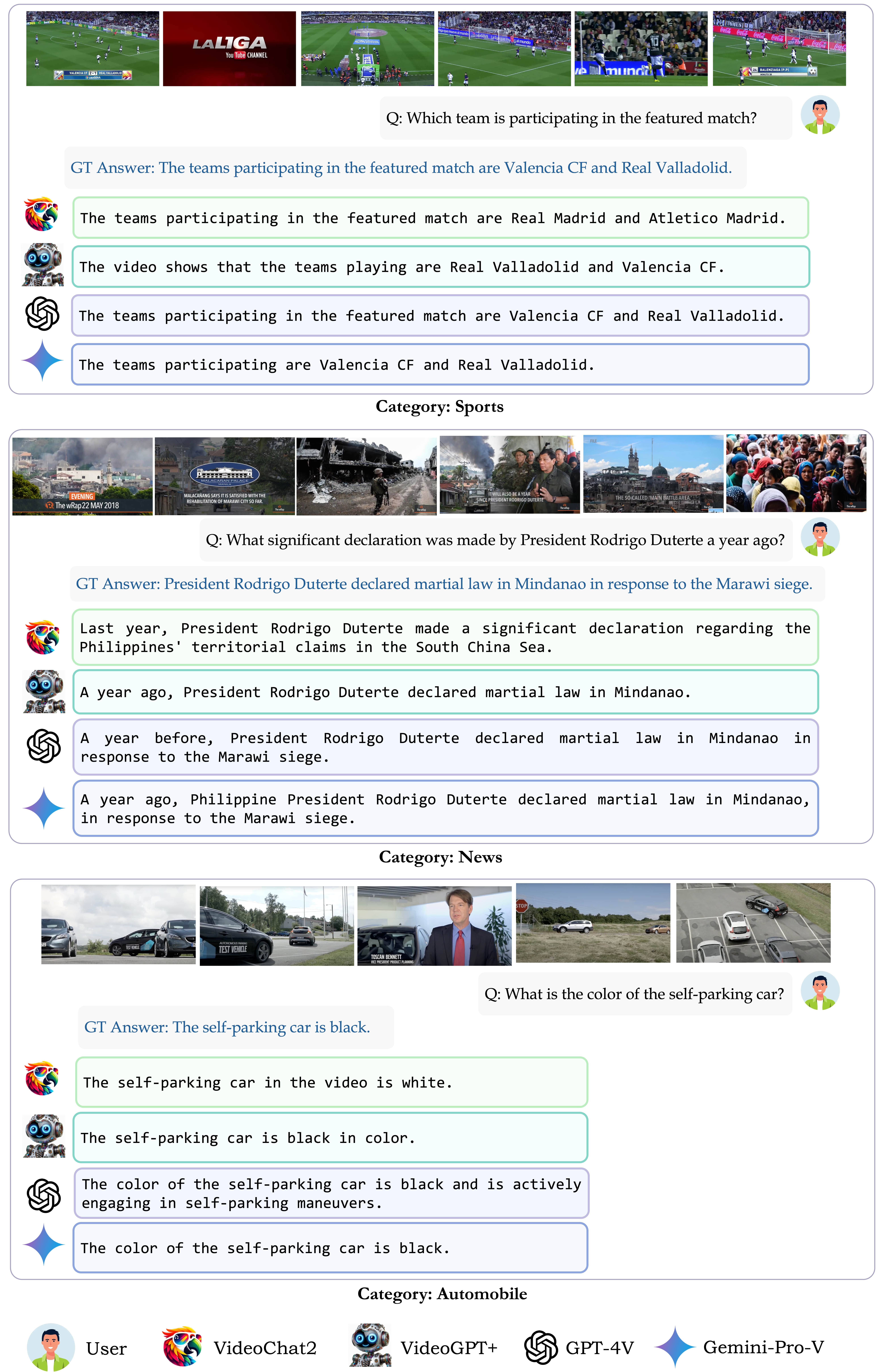}
    \caption{\small \textbf{Qualitative comparison from \ourbench of \ourmodelbold.} We show qualitative comparison of \ourmodel with VideoChat2  and propriety models GPT-4V and Gemini-1.5-Pro-V from three different categories including sports, news and automobiles videos from \ourbench.}
    \label{fig:qualitative_3}
\vspace{-1em}
\end{figure}

\section{Qualitative Results}
\label{sup:qualitative_results}
We provide a qualitative comparison of our \ourmodel with the previous state-of-the-art approach, VideoChat2~\cite{li2023mvbench}, in Fig.~\ref{fig:qualitative_1}. The example shows an advertisement video for sunscreen, where multiple scene changes are present. The video starts with a close-up view of the sunscreen, followed by a woman applying sunscreen on her hand, then applying sunscreen near a beach. The woman is then seen applying sunscreen on her arms, and finally, the video shows the key ingredients of the sunscreen and ends with the cover of the sunscreen.

As shown in Fig.~\ref{fig:qualitative_1}, our \ourmodel correctly identifies the events present in the video and provides a detailed and accurate description. On the other hand, VideoChat2 struggles to accurately capture all the events. Further, our model generates an advertisement post highlighting one of the unique features of the sunscreen shown in the video, namely that it functions as both sunscreen and moisturizer. Lastly, our \ourmodel correctly identifies the SPF value and brand name of the sunscreen, while VideoChat2 struggles to correctly identify the brand name. We present further comparison in Fig.~\ref{fig:qualitative_3}.

\section{Additional Implementation Details}
\label{sup:additional_implementation_details}
In this section, we provide additional implementation details regarding our training setup and compute requirements. All of our experiments are conducted using 8xA100 40GB GPUs. The training for VCGBench experiments takes around 12 hours to complete, while the training for MVBench experiments finishes in around 10 hours. We use the model trained for the VCGBench task to evaluate on \ourbench and zero-shot question-answering benchmarks. All of our training and evaluation codes, pretrained models and dataset will be publicly released.

\section{Additional Ablations}

\label{sup:additional_ablations}
\begin{wraptable}{r}{7.6cm}
\resizebox{0.55\textwidth}{!}{
\footnotesize
{
\begin{tabular}{lcccccc}
\toprule
\multirow{2}{*}{\textbf{Feature}} & \multicolumn{5}{c}{\textbf{VCGBench}} & \multirow{2}{*}{\textbf{Avg.}} \\ \cmidrule{2-6}
                         \textbf{Concatenation}                & \textbf{CI} & \textbf{DO} & \textbf{CU} & \textbf{TU} & \textbf{CO} &  \\ \midrule

Interleaved                                & 3.25 & 3.17 & 3.72 & 2.78 & 3.39 & 3.26 \\
\rowcolor{violet!10}  Sequential               & 3.27 & 3.18 & 3.74 & 2.83 & 3.39 & 3.28 \\
\bottomrule
\end{tabular}
}
}
\caption{\small \textbf{Ablation on Feature Concatenation Strategy.} Performance comparison between interleaved and sequential feature concatenation strategies. The sequential feature concatenation performs better.}
\label{tab:interleave_vs_sequential}
\vspace{-1em}
\end{wraptable}

\textbf{Feature concatenation strategy:} We conduct an ablation study to determine the optimal order in which image and video features should be input to the LLM. Specifically, we perform two experiments. In the first experiment, image and video features are extracted for each video segment and concatenated in an interleaved manner before sending as input to the LLM. For example, the video is divided into segments of equal size, and then the image and video features from each segment are concatenated and input to the LLM.
In the second experiment, we first place all the image features followed by all the video features. The results shown in Table~\ref{tab:interleave_vs_sequential}, indicate that the sequential design, where the image features are placed first followed by the video features, yields better performance. This can be justified by the fact that we use different visual adapters for image and video features, so interleaving the features from both modalities can create a larger distribution shift, hindering the learning process.



\begin{figure}[!h]
  \centering
    \includegraphics[width=0.97\linewidth]{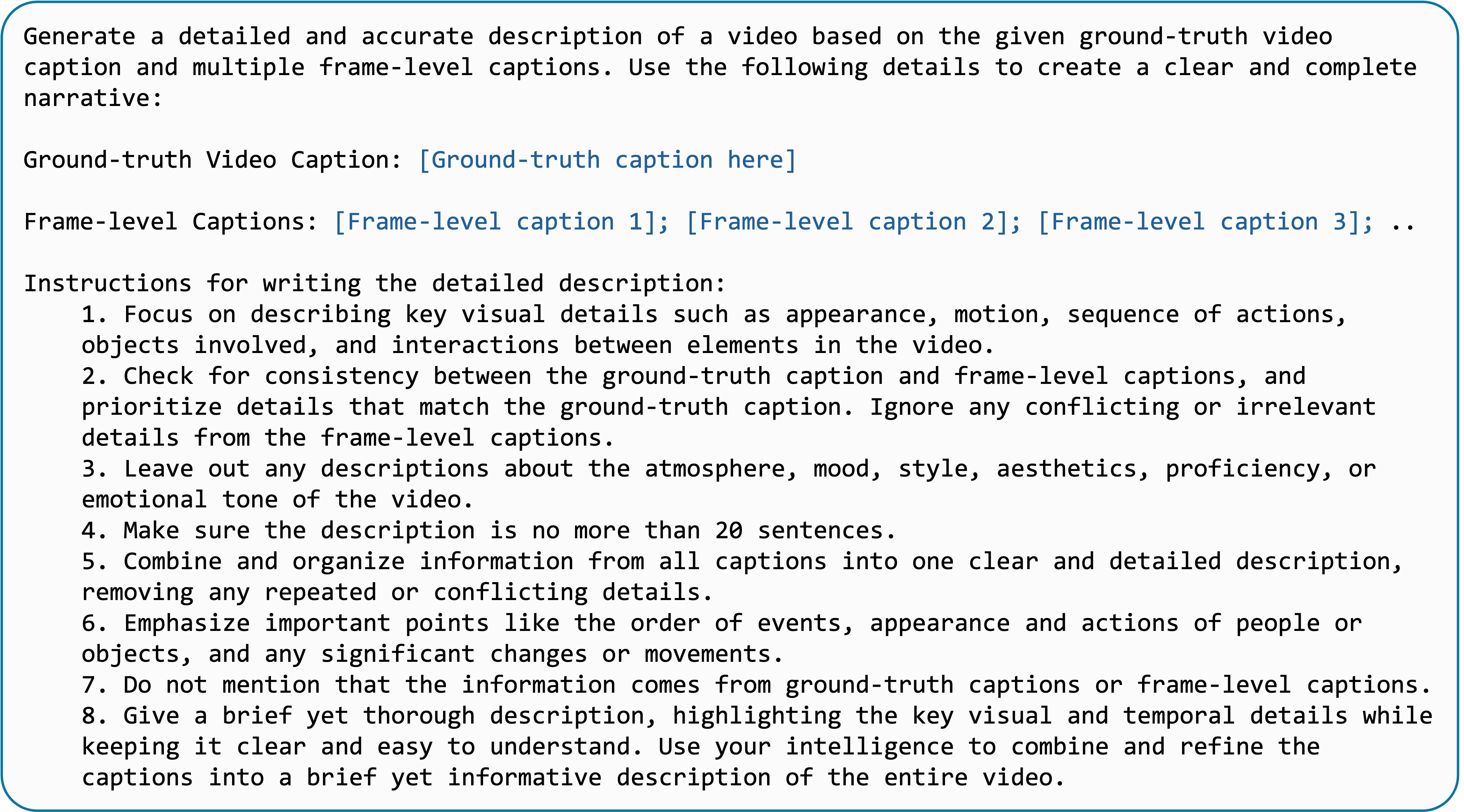}
    \caption{\small \textbf{Prompt for Dense Video Captions Generation for VCG+~112K.} We use GPT-4 to generate detailed video captions using concise ground truth and frame-level detailed captions.}
    \label{fig:prompt_1_data}
\vspace{-1em}
\end{figure}

\begin{figure}[!h]
  \centering
    \includegraphics[width=0.97\linewidth]{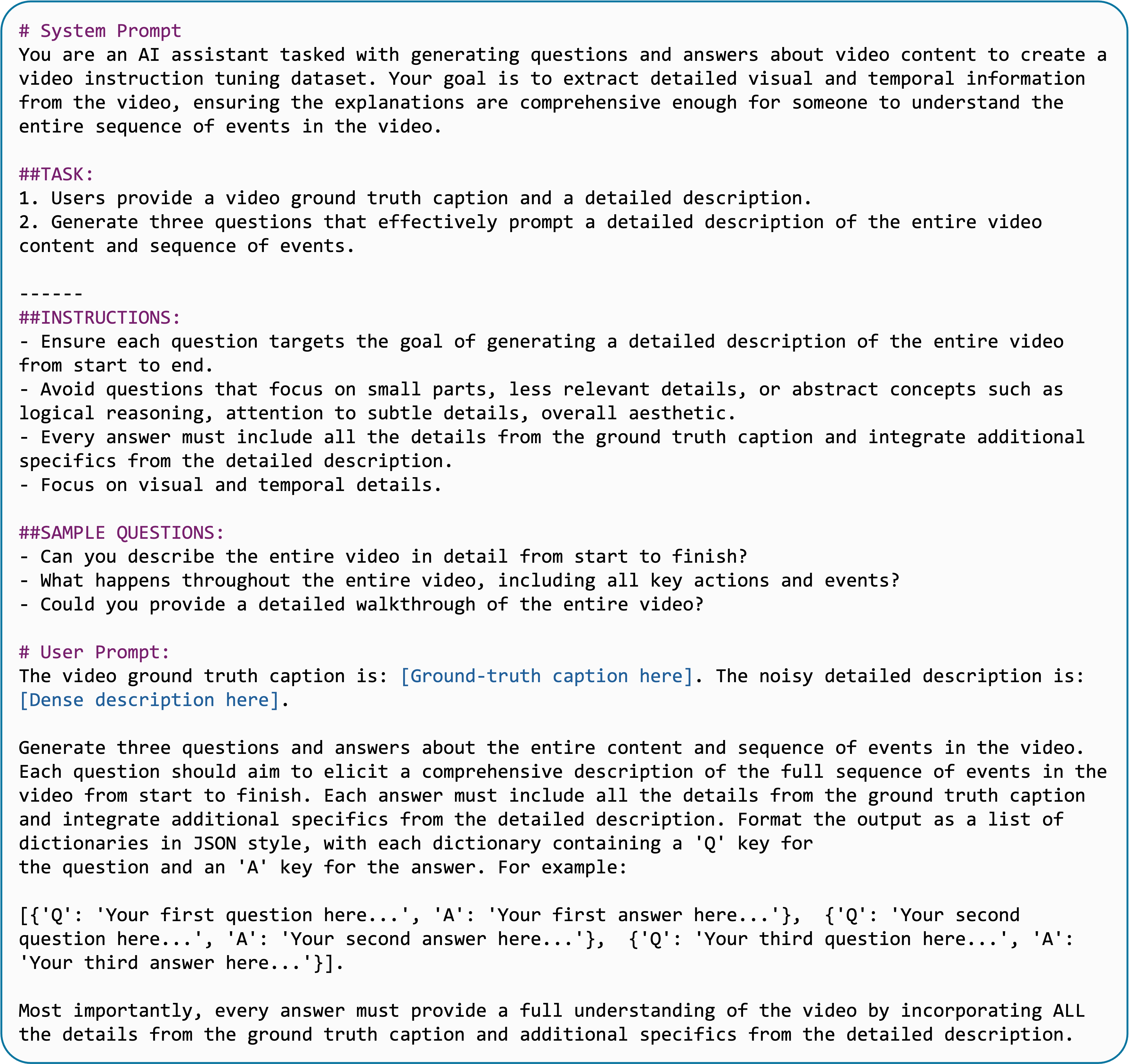}
    \caption{\small \textbf{Prompt for Question-answer generation for VCG+~112K}. We use GPT-3.5 to generate question-answer pairs for instruction tuning using the concise video ground truths and detailed video descriptions.}
    \label{fig:prompt_2_data}
\vspace{-1em}
\end{figure}

\begin{figure}[!h]
  \centering
    \includegraphics[width=1.0\linewidth]{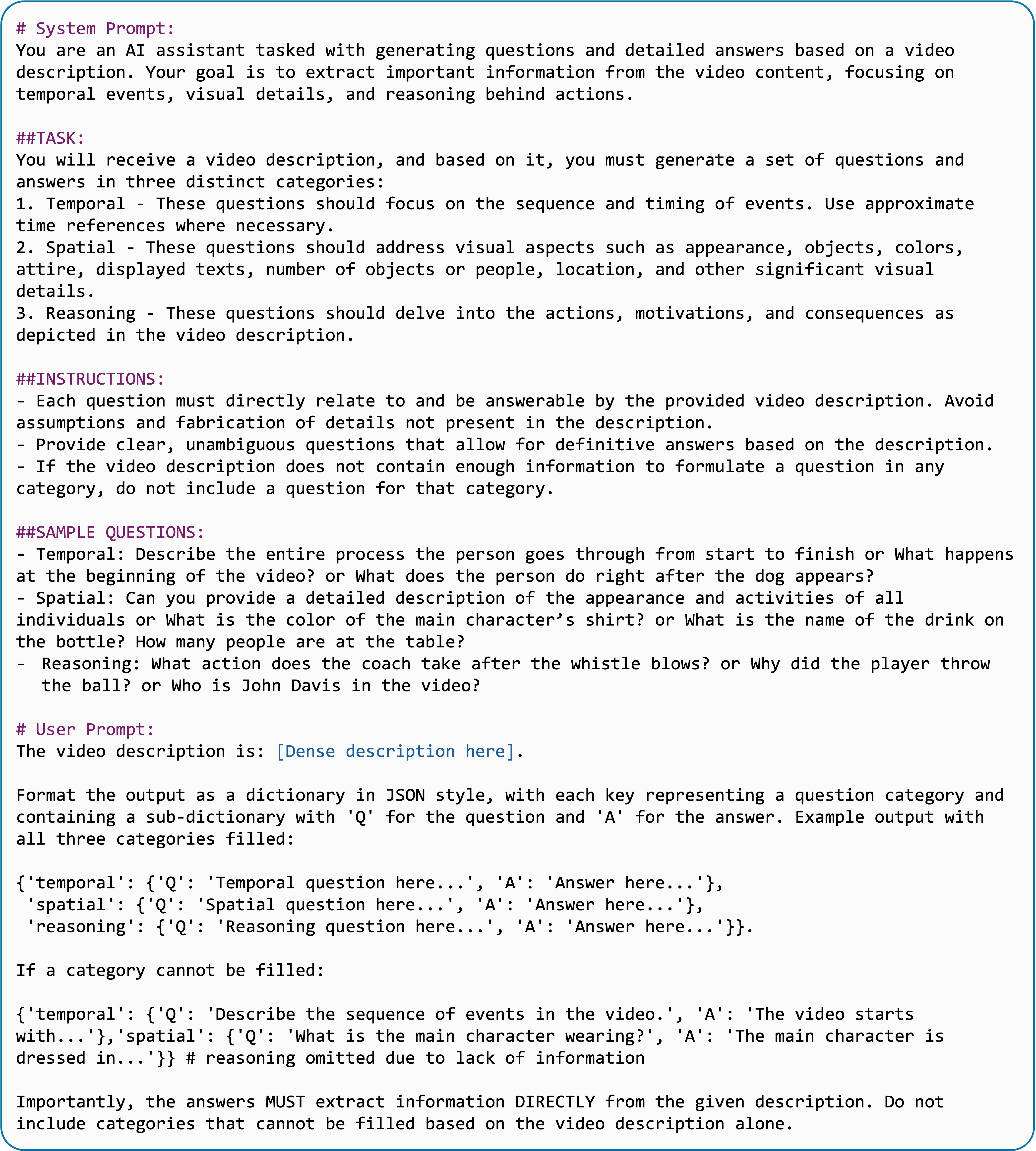}
    \caption{\small \textbf{Prompt for Question-Answer Generation for \ourbenchbold.} We use GPT-3.5 to generate temporal, spatial, and reasoning question-answer pairs.}
    \label{fig:vcgbench_diverse_dataset_generation}
\vspace{-1em}
\end{figure}

\section{GPT Prompts}
\label{sup:gpt_prompts}
In this section, we provide the GPT prompts used for the following tasks: \rom{1} Dense video description generation for \ourdata, \rom{2} Question-answer generation for \ourdata and \rom{3} Question-answer generation for \ourbench.

\textbf{Dense Video Description Generation for VCG+~112K:} To generate dense video captions, we provide GPT-4 with a concise ground truth caption of the video and detailed frame-level captions of the key-frames generated from LLaVA-v1.6~\cite{liu2024llavanext}. GPT-4 is then prompted to combine this information into a detailed caption for the entire video. As illustrated in Fig.~\ref{fig:prompt_1_data}, the prompt includes clear instructions to eliminate any conflicting information, ensuring an accurate and detailed caption.

\textbf{Question-answer generation for VCG+~112K:} After generating detailed video descriptions using GPT-4, we use GPT-3.5 to create question-answer pairs for instruction tuning.  Fig.~\ref{fig:prompt_2_data} shows the prompt to generate detailed summary question-answer pair using the ground truth caption and the dense description of the video.

\textbf{Question-Answer Generation for \ourbenchbold:} We provide prompts used to generate comprehensive question-answer pairs for \ourbench. As illustrated in Fig.~\ref{fig:vcgbench_diverse_dataset_generation}, the questions are generated in three categories: temporal, spatial, and reasoning. Similar prompts are used to generate consistency and summary questions, offering an extensive evaluation protocol for \ourbench.


\clearpage
{\small
\bibliographystyle{unsrt}
\bibliography{reference}
}
\end{document}